# Causal Modeling

John F. Lemmer
Rome Laboratory
525 Brooks Road
Griffiss AFB, NY 13441-4505
lemmer@ai.rl.af.mil

## Abstract

Causal Models are like Dependency Graphs and Belief Nets in that they provide a structure and a set of assumptions from which a joint distribution can, in principle, be computed. Unlike Dependency Graphs, Causal Models are models of hierarchical and/or parallel *processes*, rather than models of distributions (partially) known to a model builder through some sort of gestalt. As such, Causal Models are more modular, easier to build, more intuitive, and easier to understand than Dependency Graph Models. Causal Models are formally defined and Dependency Graph Models are shown to be a special case of them. Algorithms supporting inference are presented. Parsimonious methods for eliciting dependent probabilities are presented.

## 1 INTRODUCTION

Causal Models are a new knowledge representation related to Dependency Graphs, Belief Nets and Bayes Nets. Causal Models facilitate knowledge acquisition for Bayesian reasoning without relying on assumptions of independent effects or disjoint causes. Causal Models, like Dependency Graphs (Pearl 1988), are graph models, but have edges with quite different semantics. Causal Models rely on assumptions about a *domain* rather than on assumptions about a *distribution*. Causal Models are related to Probabilistic Similarity Networks (Heckerman 91) in that they allow experts to modularize their knowledge, though in a substantially different way than do Similarity Networks.

Causal Model are appropriate for domains in which many modeled events can be conceptualized as *processes* causing other events which in turn trigger other processes. Military Indications and Warning (I&W) and sensor fusion are examples of two domains having such characteristic and have motivated this work. Many domains in which diagnosis is important may also be appropriate for Causal Models, e.g. in medicine diseases can often viewed as a processes.

## 2 INFORMAL RELATION TO D-GRAPHS

In a Causal Model, a directed edge strictly means that the event at the tail of the edge *causes* the event at the head. In a Dependency-Graph (D-Graph), absence of a directed edge means that certain conditional independence assumptions are assumed true. Thus in a Causal Model an edge is an assertion about a *domain*. In a D-Graph, an edge is an assertion about a *distribution* (Pearl 1988). Pearl has suggested that in many domain/distribution pairs, conditional independence can be inferred from the lack of causal links. Our work in I&W and sensor fusion has shown such an inference cannot be made unless events unnatural to the domain experts are inserted into the model.

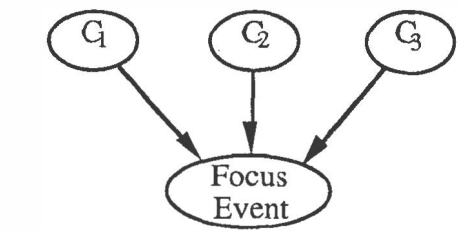

a) Traditional D-Graph Focus

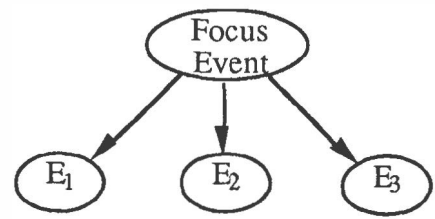

b) Additional Causal Model Focus

Figure 1:  Foci for Elicitation

In addition to different edge semantics, Causal Models allow experts to provide more and different kinds of probability information. For example, assume that a D-Graph has been constructed in which the edges *can* be interpreted as a cause effect relation. The probabilities provided by a domain expert for this model are the probabilities that various combinations of events pointing to a particular focus event will cause it. This is illustrated in Figure 1(a).

In Causal Models, however, not only can an expert provide the type of probabilities suggested by Figure 1(a), but he can also provide probabilities of the type suggested by Figure 1(b). These are the probabilities that the focus event will *cause* various combinations of the events to which it points. Thus in Causal Models, the expert provides not only the probabilities that an event will be *caused by* certain sets of events but also the probabilities that it will, in turn, *cause* other sets of events. We will show that by allowing both types of probabilities, Causal Models become extremely modular, and provide a clear semantic basis for knowledge elicitation.

In the next section we will motivate and formally define Causal Models. We will also show that every D-Graph model has an equivalent Causal Model. We will present inference algorithms for Causal Models and algorithms which substantially reduce the effort required for probability elicitation.

## 3 CAUSAL MODELS

Causal Models have been motivated by experience with Dependency Graphs. The core of the motivating experience can be understood in reference to Figure 1(a). Imagine eliciting the probability that $C_1$ alone will cause the Focus Event. If event $C_1$ is some sort of 'process', the elicited probability may seem to have two components. The first component is the probability the process will generate some sort of effect which has the *potential* of triggering the focus event. The second component is the probability that this potential triggering event will actually cause the focus event. In the mind of the expert, such a model is better expressed as in Figure 2(a). In this figure, $C_1$ has become a focus event of the type shown in Figure 1(b). In such cases there are often other internal effects of $C_1$ as suggested in Figure 2(b). Moreover, important correlations often exist among these $E_i$ caused by the $C_j$. In D-Graphs, such a correlation would be shown by an edge connecting the $E_i$. Causal Models do not show this correlation as an edge from one $E_i$ to another because such a correlation is *not* a causal relation. Causal Models represent this correlation information in the probability distribution which relates a cause to its direct effects.

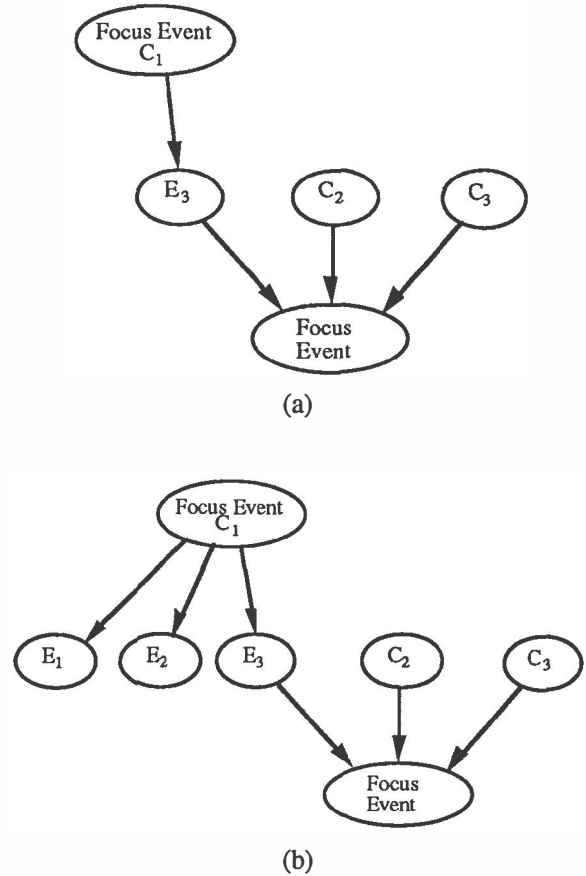

Figure 2: Causal Models

Because conditional correlations need not be inferable from edges, the graph semantics of causal models are substantially different than those of D-Graphs. In the causal model of Figure 2(b) we need not assume that $E_2$ is conditionally independent of $E_3$, given $C_1$. In a Causal Model an edge strictly means 'causes'; it does not directly relate to assertions of conditional independence.

The partial model in Figure 3 is an example of the model type shown in Figure 2(b). The focus events are shown by bold ovals. This sub-model is part of an Indications and Warnings system intended to 'diagnose' enemy intentions regarding air power. The model would be used to help infer the possible implications of observing an abnormally high number of takeoff and landings at a particular airfield[1]. Possible explanations of the high number of takeoffs are movement of supplies to the base (logistics movement), dispersal of fighter units to the base, or both. Whether one or both of these processes is in operation is key to understanding the enemy's larger intentions.

---

[1] Even in real I&W systems, such observational events are binary: either abnormally high or not.

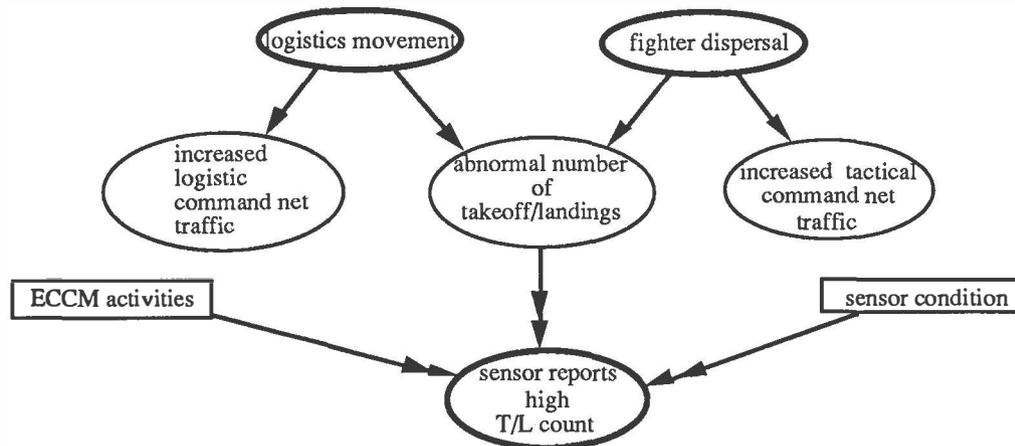

Figure 3: I&W Causal Fragment

Notice that either logistics movement or fighter dispersal, by themselves, can cause an abnormal number of takeoff and landings. If logistics movement is an active process, abnormal numbers of takeoffs and landings are likely to be correlated with increased message traffic on the logistics command nets. However, as part of a deception plan, only one or none of these effects may actually happen. Similar remarks apply to fighter dispersal and tactical command net traffic. In this example, it is reasonable to assume that logistics movement and fighter dispersal are causally independent causes (Heckerman 91) (Pearl 88) of the abnormal number of takeoffs and landings (also see below). The single arrowhead implies that causal independence is a reasonable assumption. However, it is not reasonable to assume that this abnormal number is conditionally independent of increased logistic command net traffic, given that the logistics movement process is underway. If Figure 3 were a Dependency Graph, this conditional dependence would be represented by an edge between the net traffic and takeoff/landings events. We shall show in the next section however that this approach, though known to facilitate computation, makes knowledge elicitation and model building more difficult.

In the example in Figure 3, the diagram is intended to indicate that the three causes of a *sensor report* of the abnormal count should not be regarded as casually independent. This is indicated by the double headed arrows. The Electronic Counter Counter Measures (ECCM) activities can prevent correct operation of the sensor as well as produce false alarms. The sensor condition can radically change the missed detection and false alarm rates of the sensor[2]. Thus the 'sensor' node is like a traditional D-Graph node of the type shown in Figure 1(a).

We will now provide the formal definition of a causal model, and show that D-Graphs can be considered a special case of Causal Models.

### 3.1 FORMAL DEFINITION

The graph of a Causal Model consists of nodes corresponding to two different types of events: process events and simple events. These are connected by edges meaning 'causes'. in to different ways. Without loss of generality, process events have only simple events for direct causes and direct effects. Simple events have only process events for direct causes and effects. All causes of simple events are assumed to act in a causally independent manner. In the model shown in Figure 3, the event node "abnormal number of takeoff and landings" is a simple event. The event node "Sensor reports high T/L count" is a process event.

Associated with each process node are two sets of probabilities, an 'effectual' set and a 'causal' set. The effectual set is the same distribution normally associated with an event node in a Dependency Graph, i.e., the probabilities that the (process) event will occur given the various possible combinations of its causes. As with Dependency Graphs, these conditional probabilities can be recovered from any overall distribution consistent with the Causal Model. The causal set is the probability that the process event will cause through its own 'actions' various combinations of its (simple) effects. There is no corresponding information in a D-Graph. Moreover, this distribution is not, in general, recoverable from any overall distribution consistent with the causal model.

Formally, a Causal Model, CM, is a four-tuple, $\langle P, S, C, E \rangle$. $P$ is a set of nodes containing process events; $S$ is a set of simple events. C is a set of edges, each of which leads from a process node to a simple node, i.e. $C \subset P \times S$;; $E$ is a set of edges, each of which leads from a simple event to a process event, i.e., $E \subset S \times P$. The sets, $C$ and $E$, taken together must not produce cycles in the model, i.e.,. the model with all its edges must be acyclic. A process is said to cause the set of simple events to which it is connected by elements of $C$. A process event is also said to be an effect of the simple events to which it is connected by elements of $E$, or equivalently is said to be *triggered* by the this set of events.

Two sets of probabilities are associated with each process event, its *effectual* probabilities and its *causal*

---

[2] The probability of the ECCM process being active and the probability of the various states of the sensor are presumably determined by other parts of the model not shown.

probabilities. The effectual probabilities are the probabilities normally associated with an event in Dependency Graph model. If the set $c^*$ is defined as

$$c^*(p) = \{c \mid <c, p> \in E\} \quad (1)$$

i.e., the set of simple events of which $p$ is an effect, then the effectuals are the set of probabilities

$$\Pr(p) = \{\text{pr}(p / \cap \hat{c}) \mid \hat{c} \subset c^*(p)\} \quad (2)$$

where $\cap \hat{c}$ denotes the event which is the intersection of all the events in $\hat{c}$.

The causal probabilities are defined similarly but will eventually be seen to have a rather peculiar meaning. If the set $c^\otimes$ is defined as

$$c^\otimes(p) = \{c \mid <p, c> \in C\} \quad (3)$$

then the causal probabilities of $p$ are the set

$$\tilde{\Pr}(p) = \{\tilde{\text{pr}}(\hat{c} / p) \mid \hat{c} \subset c^\otimes(p)\} \quad (4)$$

In equation (4) the probabilities are denoted by '$\tilde{\text{pr}}$' because, as will become clear, they are of a different type than the probabilities denoted by 'pr' in equation (2).

The probabilities in the set $\Pr(p)$ are interpreted as in a regular Dependency Graph, i.e.,

$$\text{pr}(p) = \sum_{\hat{c} \subseteq c^*(p)} -1^{|\hat{c}|} \text{pr}(p / \hat{c}) \, \text{pr}(\hat{c}) \quad (5)$$

where $\text{pr}(\hat{c})$ is the *marginal* probability that all the events in $\hat{c}$ occur.

However, the probabilities in the set $\tilde{\Pr}(p)$ are interpreted as causally independent causes of sets of simple events. Let $s$ be a simple event and $A(s) = \{a \mid \langle a, s \rangle \in C\}$ be the set of direct ancestors of $s$, that is the set of direct causes of $s$. If exactly the events in some subset of $A(s)$, $\hat{A}$, have occurred then, under the assumption of causal independence,

$$\text{pr}(s / \hat{A}) = 1.0 - \left[ \prod_{a \in \hat{A}} (1 - \tilde{\text{pr}}(s / a)) \right] \quad (6)$$

Equation (6) is based on the assumption that each element of $A(s)$ is capable, by itself, of causing $s$. Thus if $s$ does *not* occur, each cause must have independently (the product term) *failed* to have caused it. The probability, $\tilde{\text{pr}}(s / a)$, is the probability that $s$ is caused by $a$. It is *not* the probability that $s$ *occurs* given $a$ unless all the causes of $s$ are disjoint. This is the difference between the probability types (2) and (4).

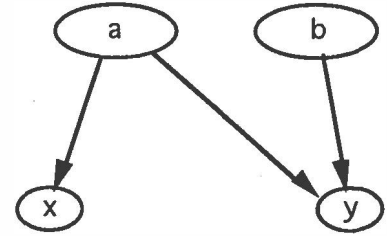

Figure 4: Co-occurring Events

If the probability of more than one event is to be calculated, application of the concept behind (6) becomes more complex. Consider computing the probability, $\text{pr}(axy)$, from the model fragment shown in Figure 4. Assume that the figure shows all the causes of the simple events, $x$ and $y$. We see from the figure that $a$ can cause both $x$ and $y$, but $b$ can cause only $y$. The required probability under the assumptions of causal independence can be computed as

$\text{pr}(axy)$
$= \{1.0 - [(1 - \tilde{\text{pr}}(x\,y\,/\,a))(1 - \tilde{\text{pr}}(x\,\bar{y}/\,a)\ \tilde{\text{pr}}(\,y/\,b))]\}\,\text{pr}(a\,b)\,(7)$
$+ \{1.0 - [\,(1 - \tilde{\text{pr}}(x\,y\,/\,a))]\}\,\text{pr}(a\,\bar{b})$

Equation (7) can be understood as follows: if $a$ and $b$ both occur, then $x$ and $y$ will not occur together only if $a$ fails to cause them both and $a$ also fails to cause just x while $b$ fails to cause y; if only $a$ occurs, then it must have failed to cause the pair. Note that from (7) $\text{pr}(ax / y) = \tilde{\text{pr}}(x\,y\,/\,a)$ if and only if $\text{pr}(a\,b) = 0$. An efficient algorithm for computing the general case is given below. However, it is based on indirect methods which provide little insight into the semantics of causal independence.

### 3.1.1 D-Graphs Are a Special Case

A simple construction exists which will convert a Dependency Graph Model into a Causal Model.

[1] Consider all events in the Dependency Graph Model to be process events and place them in the set $P$.

[2] For each edge, $\langle p_i, p_j \rangle$, in the Dependency Model, create a simple event, $s_{i,j}$ and place it in the set, $S$. Also create an edge $\langle p_i, s_{i,j} \rangle$, place it in the set $C$; create an edge $\langle s_{i,j}, p_j \rangle$ and place it in the set E.

[3] For all $p_i \in P$ set all elements of the set $\tilde{\Pr}(p_i)$ equal to 1.

This algorithm just inserts a unique simple event into each edge of a Dependency Graph. Since this event has a single ancestor and its probability of occurrence given the ancestor is unity, the probability of the original process nodes, conditioned on their ancestors, remains unchanged.

### 3.2 KNOWLEDGE ENGINEERING

The major value of Causal Modeling is the support which it provides for the knowledge engineering process. Causal Models have a theoretical basis allowing them to be built in a principled way from modular chunks of knowledge. We do not claim that Causal Models are universally applicable to all problems to which D-Graphs are applicable. But we do claim that in domains in which activation of processes are important events, Causal Models can greatly improve the knowledge engineering process.

To see how Causal Models provide principled support for modularity and hence for multiple experts begin by considering the Dependency Graph of Figure 5(a) (for the moment, without the dashed edge). Because it is a D-Graph, $a, b, c,$ and $d$ are all to be taken as process events. Interpretation of Figure 5(a) as a Dependency Graph implies that $e$ and $d$ are conditionally independent given $a$. Suppose however that this independence assumption is inappropriate. This lack of conditional independence can be expressed by adding and edge connecting $c$ and $d$ as shown by the dashed edge in Figure 5(a).

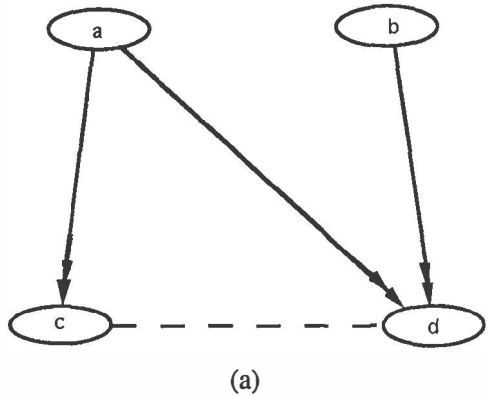

(a)

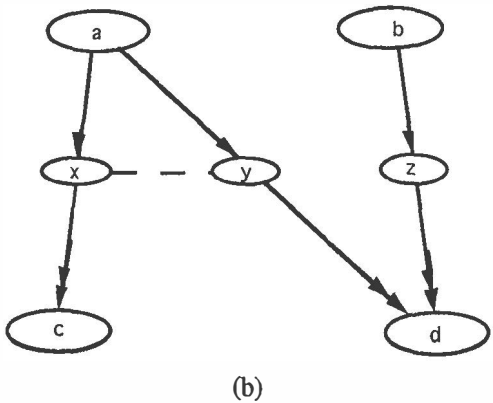

(b)

Figure 5: Correlated Effects

It may not be clear in which direction the new edge should point. The correlation implies neither that $c$ causes $d$ nor that $d$ causes $c$.

The direction assigned the edge will have significant impact on attempts at modular probability elicitation. Suppose that we have different experts for processes $c$ and $d$. If we direct the edge from $c$ to $d$, the expert in $d$ will need to provide probability estimates for $c$ being involved in the triggering of $d$, probabilities of the form $\text{pr}(d / \dot{a}\,\dot{b}\,\dot{c})^3$. The $d$ expert will probably respond that $c$ has no effect on $d$. Similar remarks would apply when the direction is from $d$ to $c$.

It is most likely that it is the expert in process $a$ who would feel that $e$ and $d$ are probably correlated given that process $a$ is active. This situation is better described by the Causal Model shown in Figure 5(b). In Figure 5(b) simple events $x$ and $y$, most likely events within process $a$, have been identified as the more specific causes of $c$ and $d$. The correlation has been associated with these events[4]. The result is that the expert in process $a$ can focus on how simple events $x$ and $y$ are correlated within process $a$, and the experts in processes $c$ and $d$ can focus on how their processes respond to these simple events if and when they occur.

When the basic domain knowledge has characteristics appropriate for a Causal Model, re-representing this knowledge as a Dependency Graph can require significant transformation of the basic knowledge. The expert is usually required to make such complex transformations mentally. As an example of the complexity, we will show how to compute the probability $\text{pr}(d/abc)$ from the parameters of the Causal Model in Figure 5(b). This probability is required by the D-Graph in Figure 5(a) if the dashed edge points from $c$ to $d$. Its value is given by $\text{pr}(d/abc) = \text{pr}(abcd)/\text{pr}(abc)$. We will compute each of the right hand factors separately.

The assumptions and structure of the model in Figure 5(b) require that $x$ occur if $c$ is to occur, and either $x$ or $y$ or both to occur if $d$ is to occur. Thus

$$\text{pr}(abcd) = \begin{bmatrix} \text{pr}(abxyzcd) \\ + \text{pr}(abxy\bar{z}cd) \\ + \text{pr}(abx\bar{y}zcd) \end{bmatrix} \text{pr}(ab) \quad (8)$$

The model further implies

---

[3] Where the dot indicates either the event or the complement of the event can appear.

[4] Process $a$ might be thought of as some sort of control process, and the simple events as the 'signals' which are sent to the processes $c$ and $d$.

$$\text{pr}(abxyzcd) = \tilde{\text{pr}}(xy/a) \times \tilde{\text{pr}}(z/b)$$
$$\times \text{pr}(c/x) \times \text{pr}(d/yz) \times \text{pr}(a)$$
$$\text{pr}(abxy\dot{z}cd) = \tilde{\text{pr}}(xy/a) \times \tilde{\text{pr}}(\dot{z}/b)$$
$$\times \text{pr}(c/x) \times \text{pr}(d/y\dot{z}) \times \text{pr}(ab)$$
$$\text{pr}(abx\bar{y}zcd) = \tilde{\text{pr}}(x\bar{y}/a) \times \tilde{\text{pr}}(z/b)$$
$$\times \text{pr}(c/x) \times \text{pr}(d/\bar{y}z) \times \text{pr}(ab)$$

Likewise

$$\text{pr}(abc) = \text{pr}(abxc)$$
$$= \text{pr}(c/x)\tilde{\text{pr}}(x/a)\text{pr}(ab) \quad (9)$$

We can now defend our claim that knowledge engineering is more modular for Causal Models than for Dependency Graphs. Causal Models are modular because the probabilities associated with Figure 5(b) can be estimated by multiple domain experts each concentrating on a single process node. It is even possible to use two experts on each node, one who understands how the process generates effects and the other who understands how the process is triggered. By contrast an expert providing the probability, $\text{pr}(d/abc)$, must, as we have just demonstrated, have knowledge of multiple nodes.

We claim superior support for Knowledge Engineering for any case where the actual domain knowledge fits the template implied by Figure 5(b). The essence of this fit is that knowledge of the domain consists of the (stochastic) understanding of processes, how processes cause simple events, and how simple events trigger processes. Dependency Graphs, on the other hand, assume the relevant knowledge is concentrated only in knowledge of process triggering. In domains we have investigated (military I&W and sensor fusion), the additional knowledge about how processes cause simple events (usually with correlation) is a key part of the domain information.

### 3.3 ALGORITHMS

To make Causal Models practical, two classes of algorithms are required, algorithms for inference, and algorithms for elicitation. The inference algorithms must convert a Causal Model into a probability distribution or a family of marginals of such a distribution, so that Bayesian inference can be performed (in any of a number of standard ways). The elicitation algorithms must make probability elicitation practically feasible.

#### 3.3.1 Inference

The following algorithm computes the joint distribution over all events in the model. It is clearly not feasible in this form except for small problems. The major reason for presenting the algorithm in this form is to present a systematic method for performing in a general way the computations analogous to those in equations (8) through (9). The description of the algorithm assumes that all events are binary. This is without loss of generality, however, because appropriate specification of the causal and effectual probabilities allows events to be declared disjoint. Disjoint events can be interpreted as discrete instances of some particular variable.

Assume that the Causal Model has a single uncaused process event, $\Omega$, which occurs with probability one[5]. Further, assume that the graph is structured so that (1) $\Omega$ is a process event, (2) process events point only to simple events, and (3) simple events point only to process events[6]. Since the graph is acyclic, we can define the 'level' of any node in the graph as the length of the *longest* path from $\Omega$ to that node. Define the depth, $D$, of a graph to be the level of the deepest node. Let $E$ be the set of all events in the model, $E = S \cup P$, and let the set

$$\text{JD}(M) = \{\text{jd}(j) \mid j \subseteq E\}$$

be a set of probabilities defined over the power set of all events in the model. If $\text{jd}(j)$ is interpreted as the probability that all the events in $j$ occur while all the events in $(E - j)$ do *not*, then $\text{JD}(M)$ is a joint distribution over all the events in the model. Given these definitions, the following algorithm computes the joint distribution implied by a Causal Model:

[1] (*Initialize; set priors for all combinations of all events caused by $\Omega$.*) Set all $\text{jd}(.) \in \text{JD}(M)$ to zero. Let $K$ [7] be the set of direct effects of $\Omega$. For all $k \subseteq K$, set $\text{jd}(k) = \tilde{\text{pr}}(k)$, $\tilde{\text{pr}}(k) \in \tilde{\text{Pr}}(\Omega)$. Set the current value of 'level' to 1.

[2] While 'level' is less than D, the model depth, repeat Steps [2.1] and [2.2],

   [2.1] (*Determine the joint probability of the process events at the current level co-occurring with other process events at this level and all events at shallower levels.*) For each process node, p at the current level, let K be the set of direct causes of event p. For each j, $j \subseteq E$ set $\text{jd}(j \cup \{p\})$ equal to $\text{jd}(j)*\text{pr}(p/j \cap K)$, $\text{pr}(p/j \cap K) \in \text{Pr}(p)$; then subtract $\text{jd}(j \cup \{p\})$ from $\text{jd}(j)$.

   [2.2] (*Determine the joint probability of the simple events at the next level co-occurring with other simple events at that level and with events at all previous levels.*) For all j, $j \subseteq E$, set $\text{jd}'(j)$ equal to 0. For each process node, p, at the

---

[5] $\Omega$ provides an easy way for specifying correlations among the 'uncaused' events in the model.

[6] This is again without loss of generality because dummy events and unitary probabilities can be inserted to trivially satisfy these conditions.

[7] The set, $K$, can be thought of as the set of simple events which are 'uncaused' within the model.

current level, let K be the set of direct effects of event p. For each j, $j \subseteq E$, and for each $k \subseteq K$ set $jd'(j \cup k)$ equal to $jd(j)*\tilde{pr}(k/p)$, $\tilde{pr}(k/p) \in \tilde{Pr}(p)$. For all j, $j \subseteq E$ set jd(j) equal to jd'(j). Increment the level by 1.

### 3.3.2 Complexity

This algorithm is clearly exponential in the number of events in the model (nodes in the graph). Step [2.2] eliminates any hope of using the graph structure to infer conditional independence. Indeed much of the motivation for this work was to capture dependencies among *effects* given the operation of various causal processes. Thus it is the nature of the knowledge we wish to model which has caused us to forego the convenience of triangulating the model into manageable cliques!

In general, stochastic sampling is probably required to make Causal Models feasible. However, some of the exponentiality of the algorithm can be tamed with two tricks. The first trick is to compute only over relevant portions of the model. The second trick is to integrate events out of the joint distribution when they no longer impact further computation. The only relevant portion of the model is the subgraph which includes the observations, causes of interest, $\Omega$, and their interconnections. Focusing only on this part of the model can greatly reduce, in any particular case, the size of the graph which must be considered. Computations of the joint distribution are impacted only by events which do not as yet have all of their directly connected effects in the distribution. Thus, if an event is neither an observation nor a cause of interest, it can be integrated from the joint distribution once all its effects are in the joint distribution. This means that at the end of each iteration of step [2], some events can be eliminated from the (growing) joint distribution.

### 3.4 ELICITATION

A Causal Model requires elicitation of both effectual and causal probabilities. Effectual probabilities are exactly the same probabilities which normally parameterize D-Graphs. An effectual probability is the probability that a particular event will occur if some particular combination of its direct potential causes occurs. A causal probability is the probability that a particular event will cause[8] a particular combination of its potential direct effects. The cardinality of the sets is exponential in both cases.

Experience has shown that experts do not like to provide the large number of parameters required to fully specify Causal and Effectual Probabilities. In D-Graphs, this exponential explosion for effectual probabilities is usually avoided by assuming causes to be either disjoint, or causally independent.

A problem introduced by Causal Models is that the very reason for their existence, to deal with non-disjoint causes and correlations among effects, makes use of the above assumptions self defeating. and elicitation of the probabilities required by the model difficult. Moreover, elicitation of causal probabilities is intrinsically more difficult than effectuals. Each element in a set of effectuals can take on any value in the closed interval [0,1]; in contrast, the sum of the elements in a set of causal probabilities must be equal to exactly one. For causal probabilities, the additivity axiom raises its ugly head[9]. We have developed methods to reduce these difficulties.

We have developed new methods, both for Causal and Effectual probabilities, to elicit reduced numbers of parameters, and to default the rest. The methods manage to achieve this without wholesale assumptions of either causal or probabilistic independence, and without assumptions of disjointedness.

### 3.4.1 Causal Probabilities

A major purpose of causal distributions in Causal Models is to model interactions among effects. Therefore elicitations based on (10) have little utility[10].

$$\text{pr}(\dot{d}, \dot{c}, \dot{b}, \dot{a},) = \text{pr}(\dot{d}) \text{pr}(\dot{c}) \text{pr}(\dot{b}) \text{pr}(\dot{a}) \quad (10)$$

Modeling a causal distribution as a Markov process does allow a distribution to be inferred from the specification of the interactions among a few events. For example, using

$$\text{pr}(\dot{d}, \dot{c}, \dot{b}, \dot{a},) = \text{pr}(\dot{d}/\dot{c}) \text{pr}(\dot{c}/\dot{b}) \text{pr}(\dot{b}/\dot{a}) \text{pr}(\dot{a}) \quad (11)$$

allows the full distribution to be estimated from seven parameters[11]. In addition, each of these parameters may take on any value in the range [0, 1], regardless of the value of the other parameters. But such models often do not correspond to the marginals with which domain experts feel most comfortable. For example, if pr($a$) is used directly in (11), pr($b$) cannot be used unless $a$ and $b$ are independent.

---

[8] *Not* co-occur with!

[9] We could proceed at this point to (re)invent Possibility Theory, but we will not!

[10] The dots indicate that the variable may stand for either an event or tits complement.

[11] E.g.
pr($a$), pr($b/a$), pr($b/\bar{a}$), pr($c/b$), pr($c/\bar{b}$), pr($d/c$), pr($d/\bar{c}$)

Our elicitation technique for causal distributions is based on eliciting sequences of marginal probabilities such as

$$\begin{aligned}&\pr(a),\ \pr(b),\ \pr(ba), \pr(c),\ \pr(ca),\\&\pr(cb),\ \pr(cba)\ \pr(d), \pr(da),\\&\pr(db), \pr(dba),\ \pr(dc),\ \pr(dca),\\&\pr(dcb),\ \pr(dcba)\end{aligned} \quad (12)$$

These are elicited in the order shown and any value may be defaulted except those involving a single event[12]. Defaulting events has results related to Markov process models as we will show later. A difficulty is that legal values for any particular probability in this series are dependent on the probability values given earlier in the series. We have developed an algorithm (Lemmer 91) which computes the legal range for an expert before he provides a value.

Sequences such as (12) allow many experts to provide probabilities for which they have good intuition. There are many other legal orderings of (12), one of which appears below in (13). By legal ordering we mean an elicitation order for which our allowable value algorithms will work correctly. Not only do such sequences allow all first order marginals to be specified, but have another interesting feature as well. In the sequence (12), consider the elicitation of pr(*dba*). When the expert must provide this value, the values for the marginals involving all the subsets of {*dba*} have already been specified. Therefore, the expert can provide, $\pr(d/ba)$, $\pr(b/da)$, or $\pr(da/b)$, etc., whichever value he feels most comfortable with. Or he can default it.

The results of defaulting values can be illustrated with the following sequence, a legal reordering of (12):

$$\begin{aligned}&\pr(a),\ \pr(b),\ \pr(ba),\ \pr(c),\ \pr(cb),\\&\pr(ca),\ \pr(cba)\ \pr(d),\ \pr(dc),\\&\pr(db),\ \pr(dcb),\ \pr(da),\ \pr(dca),\\&\pr(dba),\ \pr(dcba)\end{aligned} \quad (13)$$

If only the values in the sequence

$$\begin{aligned}&\pr(a),\ \pr(b),\ \pr(ba), \pr(c),\\&\pr(cb),\ \pr(d),\ \pr(dc)\end{aligned} \quad (14)$$

are actually provided, the same distribution results which would result from (11). The difference is that more intuitive parameters may appear in (14) for many cases.

### 3.4..2 Effectual Probabilities

---

[12] Single event probabilities could be defaulted to some value such as 0.5, but this would be essentially meaningless.

In this section we will develop a method for estimating the values of the probabilities in the set, Pr(*p*), defined in (2). The procedure will be defined in the same spirit as above. The intent is to allow interactions (i.e., correlations, dependencies, etc.) among small sets of events to specified in such a way that the interactions among larger sets can be reasonably and coherently defaulted.

In a Dependency Graph, if an event is pointed to by a set of events which cannot be assumed disjoint, it is usually necessary either to specify each of the values on the set Pr(.) or to invoke the assumption of conditional independence. Equation (15) defines the meaning of causal independence:

$$\pr(x/E,\bar{E}) = 1.0 - \prod_{e\in E}(1 - \pr(x/e)) \quad (15)$$

If the set of events, $E \cup \bar{E}$, contains all the events which point to $x$, and if the events in $E$ have occurred and those in $\bar{E}$ not, then the probability of $x$ is computed as in (15). We will generalize this in two ways: We will model how subsets of the set of events which have occurred can interact synergistically, and we will model the necessity of occurrence of subsets of the set of events which have not occurred.

Synergy can be modeled by modifying (15) as shown in (16). Note that in (16) the product iterates over subsets of $E$ rather than elements of $E$ as in (15). The quantity sy$^*$ is defined in (17) so that if all the sy (not sy$^*$) are equal to zero, (16) reduces to (15). The sy's are the synergy terms so, in this case, we say there is no synergy among the causes of $x$.

$$\pr(x/E,\bar{E}) = 1.0 - \prod_{e\subseteq E}(1 - \text{sy}^*(x/e)) \quad (16)$$

$$\text{sy}^*(x/e) = \begin{cases} pr(x/e) & if\ |e|=1 \\ sy(x/e) & otherwise \end{cases} \quad (17)$$

Equation (18) illustrates the meaning of the sy's.

$$\begin{aligned}&\pr(x/ab)\\&= [1 - (1-\pr(x/a))(1-\pr(x/b))(1-\text{sy}(x/ab))]\end{aligned} \quad (18)$$

If the value of the sy term in (18) lies in the interval (0, 1], the probability of $x$ is greater in (18) than it would be in (15). The causes, $E = \{a, b\}$ interact with positive synergy in this case. If sy is less than 0, the probability of $x$ is less than that computed from (15) and the causes have negative synergy. They inhibit each other to some extent.

To insure that (16) computes valid conditional probabilities, the sy$^*$ values must lie in the ranges shown in (19) and (20).

$$\text{sy}^*(x/e) \leq 1.0 \quad (19)$$

$$0.0 \leq \prod_{\substack{A \subseteq E \\ e \subset A}}^{\infty} (1 - \text{sy}^*(x/e)) \leq 1.0 \quad (20)$$

In (21) we modify (16) to model the necessity of possible causes which did *not* occur (i.e., are in the set $\bar{E}$). Note that if the values of all the ne's are zero, (21) reduces to (16).

$$\text{pr}(x/E, \bar{E}) = \left[1.0 - \prod_{e \subseteq E}(1 - \text{sy}^*(x/e))\right] \\ \times \prod_{\bar{e} \subseteq \bar{E}}(1 - \text{ne}(x/\bar{e})) \quad (21)$$

To see the meaning of the ne's, consider (22). Assume that $\bar{E} = \{c\}$ and that $\text{ne}(x/\bar{c})$ is equal to one. This means that the event, $c$, is absolutely necessary for $x$ to occur. Equation (22) will then evaluate to zero, no matter what the other parameters.

$$\text{pr}(x/abc) \\ = [1 - (1-\text{pr}(x/a))(1-\text{pr}(x/b))(1-\text{sy}(x/ab))] \quad (22) \\ \times (1-\text{ne}(x/c))$$

To insure that (22) computes valid condition probabilities, (19), (20), (23), and (24) must be satisfied.

$$\text{ne}(x/\bar{e}) \leq 1.0 \quad (23)$$

$$0.0 \leq \prod_{\bar{e} \subseteq \bar{E}}(1 - \text{ne}(x/\bar{e})) \leq 1.0 \quad (24)$$

Using (22), a domain expert can avoid providing most of the $2^{|E \cup \bar{E}|}$ parameters of Pr(.) while avoiding unrealistic assumptions of conditional independence.

## 4 CONCLUSION

We have presented Causal Models as a knowledge representation technique for domains which can be modeled as stochastically triggered processes which stochastically cause other triggers. An advantage of this representation is that it is modular and provides a principled way to combine knowledge acquired from a number of different experts. The representation is especially relevant in domains in which the various possible causes of particular events are not disjoint and in which events are not independent even when conditioned on their possible causes. Examples of such domains have been provided, and we have shown that Dependency Graphs are a special case of Causal Models.

We have presented methods for reducing the number of parameters which an expert, not willing to assume various types of independence, must supply to quantify correlations. These methods allow correlations (non-independences) to be specified meaningfully with a parsimonious set of numbers.

We have presented algorithms for computing the joint distribution implied by our knowledge representation. A difficulty is that, without special structure, these algorithms quickly become infeasible. 'Triangulation' algorithms for Dependency Graphs (Neapolitan 90) also rely on structure for computational feasibility, but their requirements are admittedly much less severe than the requirements for the algorithms presented here.

## 5 FURTHER RESEARCH

The author's experience has been that the representations and techniques described here facilitate the knowledge acquisition process and, when compared to traditional Dependency Graph techniques, result in better, more robust, more maintainable models. Empirical validation of the this experience would be valuable.

More efficient algorithms to support estimation of (portions of) the underlying joint distribution are needed. Just as 'triangulation' techniques take advantage of the semantics of a Dependency Graph, algorithms may be found which take advantage of the semantics of Causal Models.

### References

Heckerman, David, *Probabilistic Similarity Networks*, MIT Press, 1991.

Lemmer, John F., *SBIR Phase II: Application of Modern Mathematics to Theater Air Warfare Intelligence*, Rome Laboratory Technical Report, TR 91-111, Rome Laboratory (IRAE: Attn: Steve Farr).

Neapolitan, Richard E., *Probabilistic Reasoning in Expert Systems*, Wiley Interscience, 1990.

Pearl, Judea, *Probabilistic Reasoning in Intelligent Systems*, Morgan Kaufmann, 1988.